\pdfoutput=1
\documentclass[lettersize,journal]{IEEEtran}
\usepackage{amsmath,amsfonts,amssymb}
\usepackage{array}
\usepackage[caption=false,font=normalsize,labelfont=sf,textfont=sf]{subfig}
\usepackage{textcomp}
\usepackage{stfloats}
\usepackage{url}
\usepackage{verbatim}
\usepackage{graphicx}
\usepackage{hyperref}
\usepackage[affil-it]{authblk}
\usepackage{balance}

\usepackage{tikz}
\usepackage[edges]{forest} 
\definecolor{hidden-draw}{RGB}{106,142,189} 
\definecolor{hidden-blue}{RGB}{194,232,247} 
\definecolor{hidden-orange}{RGB}{217, 232, 252} 
\definecolor{trainingColor}{RGB}{180, 0, 0} 
\definecolor{finetuningColor}{RGB}{0, 102, 204} 
\definecolor{noFinetuningColor}{RGB}{0, 128, 0} 
\definecolor{knowledgeColor}{RGB}{128, 0, 128} 
\definecolor{peftColor}{RGB}{255, 140, 0}   \newcommand{\trainSym}{\textcolor{trainingColor}{$\bullet$}}
\newcommand{\finetuneSym}{\textcolor{finetuningColor}{\tiny$\blacksquare$}}
\newcommand{\noFinetuneSym}{\textcolor{noFinetuningColor}{$\blacktriangle$}}
\newcommand{\knowledgeSym}{\textcolor{knowledgeColor}{$\blacklozenge$}}
\newcommand{\peftSym}{\textcolor{peftColor}{$\bigstar$}}  

\usepackage{caption}

\usepackage{booktabs}
\usepackage{mathabx}
\usepackage{algorithm}
\usepackage[noend]{algpseudocode}
\usepackage{multirow}
\usepackage{wrapfig}

\begin{document}
\renewcommand{\thefootnote}{\fnsymbol{footnote}}
\title{Speculative Decoding and Beyond: An In-Depth Survey of Techniques}

\author{
  \textbf{Yunhai Hu}\textsuperscript{1}\textsuperscript{*},
  \textbf{Zining Liu}\textsuperscript{2}\textsuperscript{*},
  \textbf{Zhenyuan Dong}\textsuperscript{1}\textsuperscript{*},
  \textbf{Tianfan Peng}\textsuperscript{1,3}\textsuperscript{*},
  \textbf{Bradley McDanel}\textsuperscript{4},
  \textbf{Sai Qian Zhang}\textsuperscript{1}\textsuperscript{†}
  \\
  \textsuperscript{1}New York University,
  \textsuperscript{2}University of Pennsylvania,
  \textsuperscript{3}Shenzhen Institute of Information Technology,
  \textsuperscript{4}Franklin and Marshall College
  \\
  \texttt{\{yunhai.hu, zd2362, sai.zhang\}@nyu.edu}
  \texttt{zliu0@seas.upenn.edu}
  \texttt{tianfanpeng@gmail.com}
  \texttt{bmcdanel@fandm.edu}
}

\maketitle
\footnotetext[1]{Equal contributions.}
\footnotetext[2]{Corresponding author.}

\begin{abstract}
Sequential dependencies present a fundamental bottleneck in deploying large-scale autoregressive models, particularly for real-time applications. While traditional optimization approaches like pruning and quantization often compromise model quality, recent advances in generation-refinement frameworks demonstrate that this trade-off can be significantly mitigated. 

This survey presents a comprehensive taxonomy of generation-refinement frameworks, analyzing methods across autoregressive sequence tasks. We categorize methods based on their generation strategies (from simple n-gram prediction to sophisticated draft models) and refinement mechanisms (including single-pass verification and iterative approaches). Through systematic analysis of both algorithmic innovations and system-level implementations, we examine deployment strategies across computing environments and explore applications spanning text, images, and speech generation. This systematic examination of both theoretical frameworks and practical implementations provides a foundation for future research in efficient autoregressive decoding.
\end{abstract}
\section{Introduction}
\label{sec:introduction}

Large Models (LMs) have demonstrated remarkable capabilities across diverse domains, from text generation~\cite{brown2020language,45-llm-qa,34-llama} and translation~\cite{zhu2023multilingual,46-language-translation,huang2023towards} to image synthesis~\cite{ho2020denoising,yang2023diffusion,tian2024visual} and video generation~\cite{SORA,wu2023tune,opensora}. However, these models face a critical challenge: their inherently sequential nature creates significant latency bottlenecks, particularly for real-time applications. While traditional optimization approaches like quantization and pruning often compromise model quality for speed, recent research has focused on maintaining output quality while breaking sequential dependencies through novel algorithmic and system-level innovations.

Generation-refinement frameworks have emerged as a promising family of solutions that directly address these sequential bottlenecks. These approaches encompass a range of methods, from speculative decoding with draft models to iterative refinement techniques inspired by numerical optimization. The common thread among these approaches is their division of the generation process into two phases: an initial generation step that produces draft tokens in parallel, followed by a refinement step that ensures output quality. 

The implementation of these frameworks presents unique system-level challenges across different deployment scenarios. Edge devices require careful optimization of memory usage and computation patterns~\cite{svirschevski2024specexec,xu2024edgellm}, while distributed systems must manage complex communication patterns and load balancing. These system-level considerations have driven innovations in areas like kernel design, hardware acceleration, and batch processing optimization, significantly influencing both algorithmic choices and practical performance.

This survey synthesizes research across these approaches, examining both algorithmic innovations and their system implementations. We present a systematic taxonomy of generation-refinement methods, analyze deployment strategies across computing environments, and explore applications spanning text, images~\cite{wang2024continuous,jang2024lantern}, and speech~\cite{li2024fast,raj2024faster}. Our primary contributions include comprehensive analysis of system-level implementations and optimizations, detailed examination of applications across modalities, and identification of key research challenges in efficient neural sequence generation.

\section{The Sequential Bottleneck in Large Model Inference}
\label{sec:sequential_bottleneck}

\subsection{Understanding Sequential Dependencies}
\label{sec:sequential_dependencies}

\begin{figure*}
    \centering
    \includegraphics[width=0.9\linewidth]{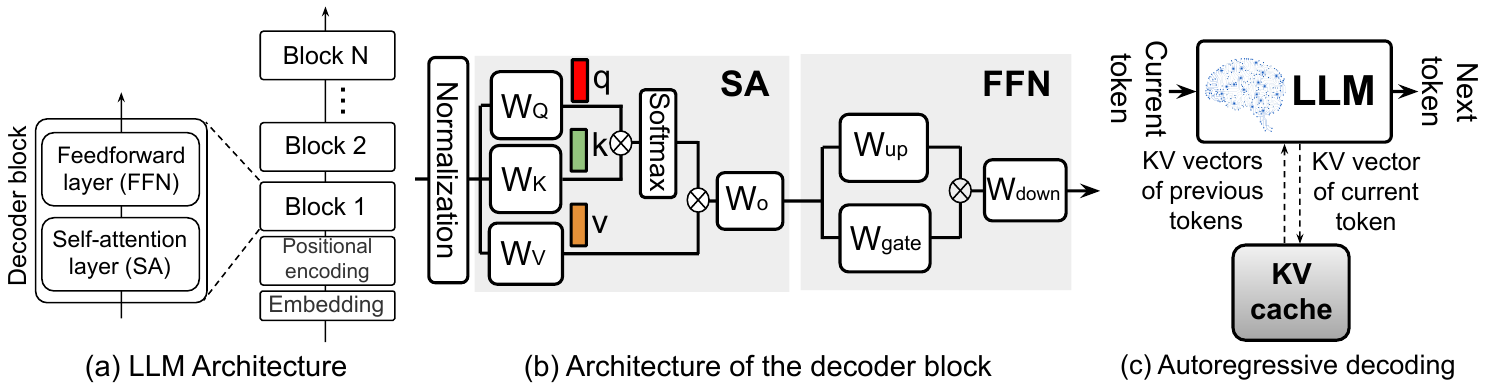}
    \caption{(a) The Llama architecture consists of stacked transformer decoder blocks. (b) Each decoder block contains a self-attention (SA) block and feedforward (FFN) block. (c) During the decoding stage, tokens are generated auto-regressively.}
    \label{fig:architech}
\end{figure*}

Modern LLMs, such as the Llama series~\cite{touvron2023llama,touvron2023llama2,dubey2024llama} and the GPT series~\cite{radford2019language,brown2020language}, are built on transformer architectures consisting of stacked decoder blocks. As shown in Figure~\ref{fig:architech}(a), each decoder block contains two fundamental components: a Self-Attention (SA) block and a feed-forward network (FFN). During execution, the input of the SA block is first multiplied with three weight matrices $W_{Q}$, $W_{K}$, and $W_{V}$, yielding the outputs termed query ($q$), key ($k$), and value ($v$), respectively.

The computation flow, detailed in Figure~\ref{fig:architech}(b), shows how query and key vectors compute attention scores through matrix multiplication. After softmax normalization, these scores weight the value vectors, producing the SA output through a weighted sum and residual connection. This SA output feeds into the FFN, typically implemented as either a standard MLP~\cite{radford2018improving, radford2019language} or gated MLP~\cite{liu2021pay, touvron2023llama,touvron2023llama2}, with multiple fully connected layers and activation functions like GeLU~\cite{hendrycks2016gaussian} or SiLU~\cite{elfwing2018sigmoid}.

The core challenge emerges during inference, which consists of two main phases: prefill and decoding. While the prefill phase can process input sequences in parallel, the decoding phase introduces a critical bottleneck. As shown in Figure~\ref{fig:architech}(c), the model must predict each token sequentially, using both current and previous token information through their Key and Value (KV) vectors. These KV vectors are cached for subsequent predictions, leading to significant memory access latency as the sequence length grows.

\subsection{Breaking Sequential Dependencies}
\label{sec:breaking_dependencies}

Traditional approaches to accelerating LM inference have focused on reducing computational costs through model compression, knowledge distillation, and architectural optimizations. However, these methods primarily address individual computation costs rather than the fundamental sequential dependency that requires each token to wait for all previous tokens.

\begin{figure}
    \centering
    \includegraphics[width=0.85\linewidth]{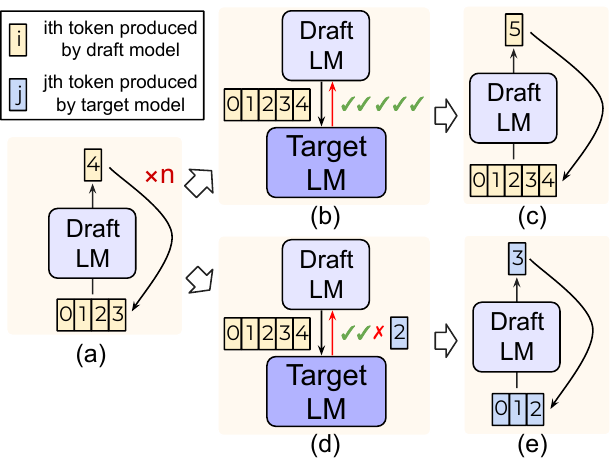}
    \caption{Illustration of speculative decoding workflow.}
    \label{fig:sd_intro}
\end{figure}

Speculative decoding (SD)~\cite{stern2018blockwise} has emerged as a promising solution that directly targets this sequential bottleneck. As illustrated in Figure~\ref{fig:sd_intro}, this approach introduces a two-phase process where a smaller, faster \textit{draft model} first predicts multiple tokens in parallel, followed by verification using the target model. The draft model enables parallel token generation, breaking away from traditional token-by-token generation, while the target model's verification step maintains output quality through accept/reject decisions.

This strategy has proven particularly valuable for real-time applications like interactive dialogue systems, where response latency directly impacts user experience. The verification mechanism provides a crucial balance between generation speed and output quality, accepting correct predictions to maintain throughput while falling back to sequential generation when necessary to preserve accuracy.

While SD represents one successful approach to breaking sequential dependencies in autoregressive (AR) models, it belongs to a broader family of \textit{generation-refinement} methods. The following sections present a systematic taxonomy of these approaches, examining how different techniques balance the trade-offs between generation parallelism and output quality.
\section{A Taxonomy for Generation and Refinement Frameworks}
\label{sec:sd_taxonomy}

\begin{figure}
    \centering
    \includegraphics[width=0.9\linewidth]{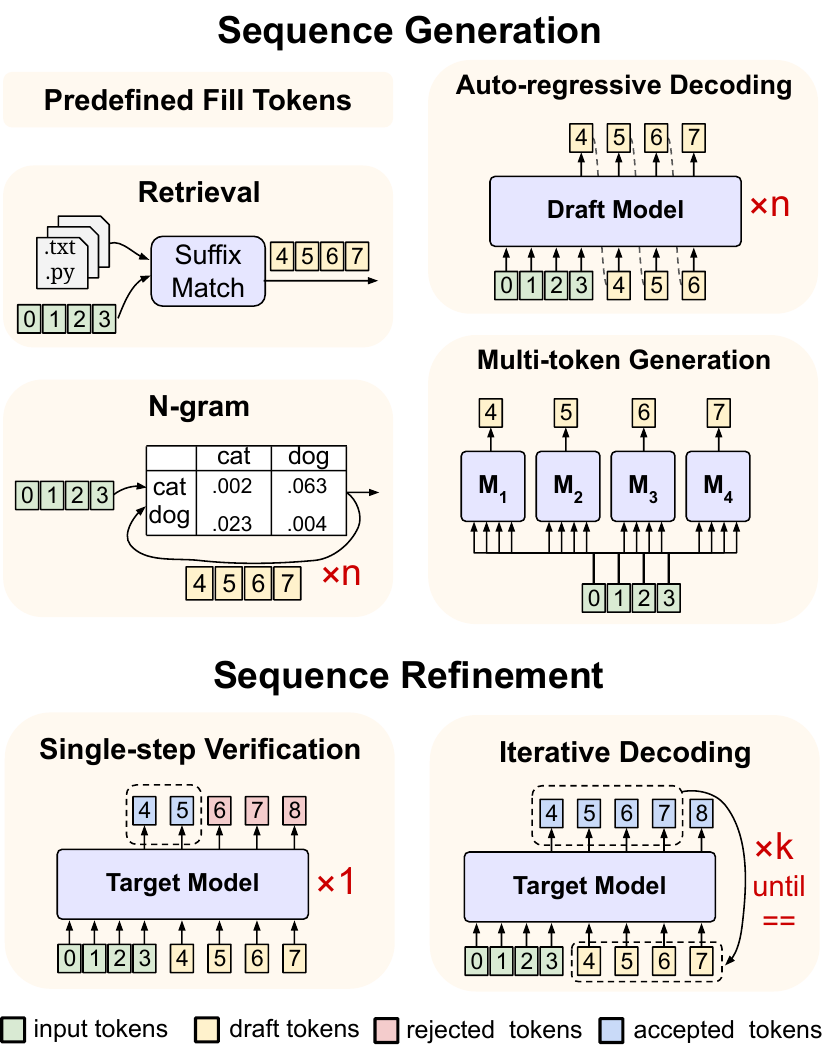}
    \caption{A taxonomy of generation-refinement frameworks, showing two phases: (1) Generation of draft tokens through various methods and (2) Refinement through verification strategies.}
    \label{fig:overview}
\end{figure}

To systematically analyze approaches for breaking sequential dependencies in large models, we propose a unified taxonomy that categorizes methods based on their generation and refinement strategies. As shown in Figure~\ref{fig:overview}, our taxonomy decomposes these frameworks into two fundamental phases: \textit{Sequence Generation} and \textit{Sequence Refinement}. This decomposition not only encompasses traditional SD approaches but also captures a broader range of emerging methods that trade off between generation parallelism and output quality.

The sequence generation phase focuses on different strategies for producing draft tokens more efficiently than conventional auto-regressive decoding using a single larger model. These strategies range from simple approaches like random token sampling (used in conjunction with iterative decoding) to more sophisticated methods like retrieval-based generation and draft model prediction. Each generation method offers trade-offs in terms of computational cost and prediction quality. The sequence refinement phase then determines how these candidates are processed - either accepting them directly (with possible poorer quality), verifying a subset of tokens in a single pass, or refining the draft tokens through multiple iterations until convergence.

\tikzstyle{my-box}=[
 rectangle,
 draw=hidden-draw,
 rounded corners,
 text opacity=1,
 minimum height=1.5em,
 minimum width=5em,
 inner sep=2pt,
 align=center,
 fill opacity=.5,
 ]
 \tikzstyle{leaf}=[my-box, minimum height=1.5em,
 fill=hidden-orange!60, text=black, align=left,font=\scriptsize,
 inner xsep=2pt,
 inner ysep=4pt,
 ]

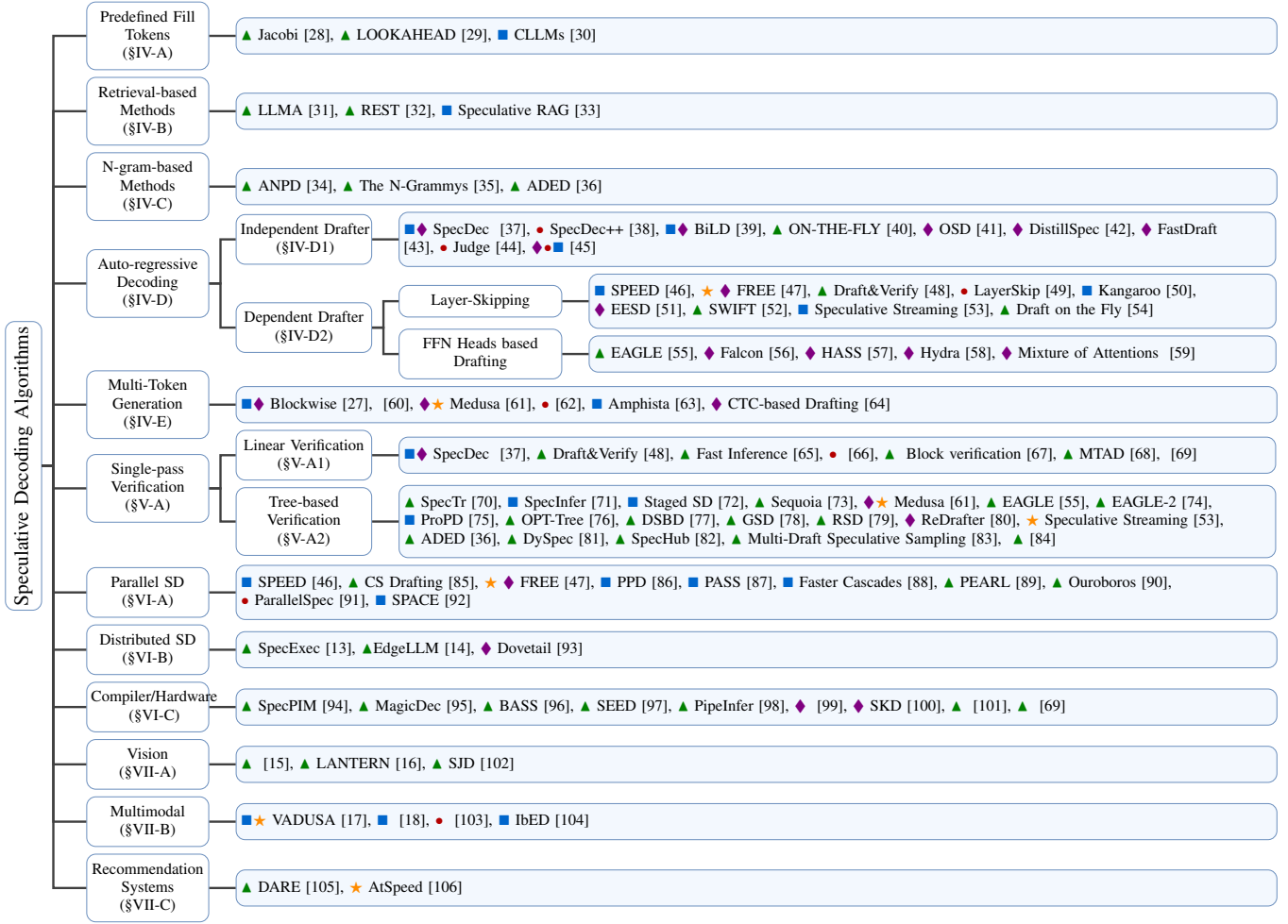
\begin{figure*}[t]
	\centering
	\resizebox{\textwidth}{!}{
		\begin{forest}
			forked edges,
			for tree={
				grow=east,
				reversed=true,
				anchor=base west,
				parent anchor=east,
				child anchor=west,
                node options={align=center},
                align = center,
				base=left,
				font=\small,
				rectangle,
				draw=hidden-draw,
				rounded corners,
				minimum width=4em,
				edge+={darkgray, line width=1pt},
				s sep=3pt,
				inner xsep=2pt,
				inner ysep=3pt,
				ver/.style={rotate=90, child anchor=north, parent anchor=south, anchor=center},
			},
			where level=1{text width=5.0em,font=\scriptsize}{},
			where level=2{text width=5.6em,font=\scriptsize}{},
			where level=3{text width=6.8em,font=\scriptsize}{},
			[
			Speculative Decoding Algorithms, ver
                [
			  Predefined Fill \\ Tokens \\ (\S\ref{sec:predefined_fill_tokens})
			[ 
                \noFinetuneSym~Jacobi~{\cite{santilli2023accelerating},}  
                \noFinetuneSym~LOOKAHEAD~{\cite{fu2024break},}  
                \finetuneSym~CLLMs~{\cite{kou2024cllms}}
                , leaf, text width=45.67em, align = left
			]
			]
                [
			  Retrieval-based \\ Methods \\      (\S\ref{sec:retrieval_based_methods})
			[ 
                \noFinetuneSym~LLMA~{\cite{yang2023inference},}
                \noFinetuneSym~REST~{\cite{he2023rest},}
                \finetuneSym~Speculative RAG~{\cite{wang2024speculative}}
                , leaf, text width=45.67em, align = left
			]
			]
                [
			  N-gram-based \\ Methods \\ (\S\ref{sec:ngram_methods})
			[ 
                \noFinetuneSym~ANPD~{\cite{ou2024lossless},}
                \noFinetuneSym~The N-Grammys~{\cite{stewart2024n},}
                \noFinetuneSym~ADED~{\cite{liu2024adaptive}}
                , leaf, text width=45.67em, align = left
			]
			]
			[
			Auto-regressive \\ Decoding \\
                (\S\ref{sec:auto_regressive})
			[
			  Independent Drafter \\ (\S\ref{sec:independent_drafter})
			[
                \finetuneSym\knowledgeSym~SpecDec~
                {\cite{xia2023speculative},}
                \trainSym~SpecDec++~{\cite{huang2024specdec++},}
                \finetuneSym\knowledgeSym~BiLD~{\cite{kim2024speculative},}
                \noFinetuneSym~ON-THE-FLY~{\cite{liu2025a},}
                \knowledgeSym~OSD~{\cite{liu2023online},} 
			\knowledgeSym~DistillSpec~{\cite{zhou2023distillspec},} 
                \knowledgeSym~FastDraft~ \\{\cite{zafrir2024fastdraft},} 
                \trainSym~Judge~{\cite{bachmann2025judge},} 
                \knowledgeSym\trainSym\finetuneSym~{\cite{Liu2025}} 
			, leaf, text width=38.4em, align = left
			]
			]
			[
			  Dependent Drafter \\ (\S\ref{sec:dependent_drafter})
                [
			  Layer-Skipping 
			[
                \finetuneSym~SPEED~{\cite{hooper2023speed},} 
                \peftSym~\knowledgeSym~FREE~{\cite{bae2023fast},} 
			\noFinetuneSym~Draft\&Verify~{\cite{zhang2023draft},} 
                \trainSym~LayerSkip~{\cite{elhoushi2024layer},} 
                \finetuneSym~Kangaroo~{\cite{liu2024kangaroo},} \\ 
                \knowledgeSym~EESD~{\cite{liu2024speculative},}
                \noFinetuneSym~SWIFT~{\cite{xia2024swift},} 
                \finetuneSym~Speculative Streaming~{\cite{bhendawade2024speculative},} 
                \noFinetuneSym~Draft on the Fly~{\cite{metel2024draft}}
			, leaf, text width=30em, align = left
			]
			]
			[
			  FFN Heads based \\ Drafting    
			[
                \noFinetuneSym~EAGLE~{\cite{li2024eagle},}
                \knowledgeSym~Falcon~{\cite{gao2024falcon},}
                \knowledgeSym~HASS~{\cite{zhang2024learning},} 
                \knowledgeSym~Hydra~{\cite{ankner2024hydra},} 
                \knowledgeSym~Mixture of Attentions ~{\cite{zimmer2024mixture}} 
			, leaf, text width=30em, align = left
			]
			]
			]
			]
                [
			  Multi-Token \\ Generation \\ (\S\ref{sec:multi_token_pre})
			[
                \finetuneSym\knowledgeSym~Blockwise~{\cite{stern2018blockwise},}
                 ~{\cite{kim2024accelerating},}
			\knowledgeSym\peftSym~Medusa~{\cite{cai2024medusa},} 
                \trainSym~{\cite{gloeckle2024better},}
                \finetuneSym~Amphista~{\cite{li2024amphista},} 
                \knowledgeSym~CTC-based Drafting~{\cite{wen2024speculative}} 
                , leaf, text width=45.67em, align = left
			]
			]
                [
			  Single-pass \\ Verification \\ (\S\ref{sec:refine:singlepass})
                [
			Linear Verification \\(\S\ref{sec:refine:linear})
			[
			    \finetuneSym\knowledgeSym~SpecDec~
                {\cite{xia2023speculative},}
                \noFinetuneSym~Draft\&Verify~{\cite{zhang2023draft},} 
                \noFinetuneSym~Fast Inference{~\cite{leviathan2023fast},}  
                \trainSym~{~\cite{chen2023accelerating},} 
                \noFinetuneSym~ Block verification~{\cite{sun2025block},}
                \noFinetuneSym~MTAD~{\cite{qin2024optimized},}
                ~{\cite{yin2024theoretical}} 
			, leaf, text width=38.4em, align = left
			]
                ]
                [
			Tree-based \\ Verification \\ (\S\ref{sec:refine:tree_based})
			[
			\noFinetuneSym~SpecTr~{\cite{sun2024spectr},} 
                \finetuneSym~SpecInfer~{\cite{miao2023specinfer},}  
                \finetuneSym~Staged SD~{\cite{spector2023accelerating},} 
                \noFinetuneSym~Sequoia~{\cite{chen2024sequoia},} 
                \knowledgeSym\peftSym~Medusa~{\cite{cai2024medusa},} 
                \noFinetuneSym~EAGLE~{\cite{li2024eagle},}
                \noFinetuneSym~EAGLE-2~{\cite{li2024eagle2fasterinferencelanguage},} \\
                \finetuneSym~ProPD~{\cite{zhong2024propd},}
                \noFinetuneSym~OPT-Tree~{\cite{wang2024opt},} 
                \noFinetuneSym~DSBD~{\cite{qin2024dynamic},}
                \noFinetuneSym~GSD~{\cite{gong2024graph},} 
                \noFinetuneSym~RSD~{\cite{jeon2024recursive},} 
                \knowledgeSym~ReDrafter~{\cite{cheng2024recurrent},}
                \peftSym~Speculative Streaming~{\cite{bhendawade2024speculative},} \\
                \noFinetuneSym~ADED~{\cite{liu2024adaptive},}
                \noFinetuneSym~DySpec~{\cite{xiong2024dyspec},}
                \noFinetuneSym~SpecHub~{\cite{sun2024spechub},} 
                \noFinetuneSym~Multi-Draft Speculative Sampling~{\cite{khisti2024multi},} 
                ~\noFinetuneSym~{\cite{hu2025towards}}
			, leaf, text width=38.4em, align = left
			]
                ]
			]
                [
			  Parallel SD \\ (\S\ref{sec:parallel-sd})
			[
                 \finetuneSym~SPEED~{\cite{hooper2023speed},} 
                \noFinetuneSym~CS Drafting~{\cite{chen2023cascade},}  
                \peftSym~\knowledgeSym~FREE~{\cite{bae2023fast},}
                \finetuneSym~PPD~{\cite{yang2023predictive},} 
                \finetuneSym~PASS~{\cite{monea2023pass},} 
                \finetuneSym~Faster Cascades~{\cite{narasimhan2024faster},} 
                \noFinetuneSym~PEARL~{\cite{liu2024parallel},} 
                \noFinetuneSym~Ouroboros~{\cite{zhao-etal-2024-ouroboros},} \\
                \trainSym~ParallelSpec~{\cite{xiao2024parallelspec},}
                \finetuneSym~SPACE~{\cite{yi-etal-2024-generation}}
                , leaf, text width=45.67em, align = left
			]
			]
                [
                Distributed  SD \\ (\S\ref{sec:edge-sd})
                [
                \noFinetuneSym~{SpecExec~\cite{svirschevski2024specexec},}
                \noFinetuneSym{EdgeLLM~\cite{xu2024edgellm},}
                \knowledgeSym~{Dovetail~\cite{zhang2024dovetail}}
                , leaf, text width=45.67em, align = left
                ]
                ]
                [
                Compiler/Hardware \\ (\S\ref{sec:batch-sd})
                [
                \noFinetuneSym~{SpecPIM~\cite{li2024specpim},}
                \noFinetuneSym~{MagicDec~\cite{chen2024magicdec},}
                \noFinetuneSym~{BASS~\cite{qian2024bass},}
                \noFinetuneSym~{SEED~\cite{wang2024seed},}
                \noFinetuneSym~{PipeInfer~\cite{butler2024pipeinfer},} 
                \knowledgeSym~{~\cite{wang2024mamba},}
                \knowledgeSym~{SKD~\cite{xu2024speculative},}
                \noFinetuneSym~{~\cite{wagner2024optimized},}
                \noFinetuneSym~{~\cite{yin2024theoretical}}
                , leaf, text width=45.67em, align = left
                ]
                ]
                [
                Vision \\ (\S\ref{sec:AR_visual})
                [
                \noFinetuneSym~{~\cite{wang2024continuous},}
                \noFinetuneSym~{LANTERN~\cite{jang2024lantern},}
                \noFinetuneSym~{SJD~\cite{teng2024accelerating}}
                , leaf, text width=45.67em, align = left
                ]
                ]
                [
                Multimodal \\ (\S\ref{sec:mutimodel})
                [
                \finetuneSym\peftSym~{VADUSA~\cite{li2024fast},}
                \finetuneSym~{~\cite{raj2024faster},}
                \trainSym~{~\cite{gagrani2024speculative},}
                \finetuneSym~{IbED~\cite{leebatch}}
                , leaf, text width=45.67em, align = left
                ]
                ]
                [
                Recommendation\\Systems \\ (\S\ref{sec:SR_apps})
                [
                \noFinetuneSym~{DARE~\cite{xi2024decoding},}
                \peftSym~{AtSpeed~\cite{lin2024efficient}}
                , leaf, text width=45.67em, align = left
                ]
                ]
			]
		\end{forest}
  }

\caption{Taxonomy of Speculative Decoding Algorithms. Symbols indicate implementation approach: \noFinetuneSym~Direct application (no training required), \trainSym~Full model training from scratch, \finetuneSym~Model fine-tuning, \peftSym~Parameter-efficient fine-tuning (PEFT), \knowledgeSym~Knowledge distillation from target model.}
\label{Speculative_decoding_algorithm}
\end{figure*}

\section{Sequence Generation Methods}
\label{sec:generation}

\subsection{Predefined Fill Tokens}
\label{sec:predefined_fill_tokens}
The simplest approach uses random initialization or predefined tokens (e.g., \texttt{PAD}). While computationally free, these methods provide poor initialization points, requiring multiple refinement iterations as discussed in Section~\ref{sec:refine:iterative}.

\subsection{Retrieval-based Methods}
\label{sec:retrieval_based_methods}
LLMA ~\cite{yang2023inference} first proposed exploiting overlaps between LLM outputs and reference documents to accelerate inference through parallel token verification while maintaining identical generation results. In retrieval-based approaches, REST~\cite{he2023rest} replaces smaller language models with exact suffix matching from a datastore to generate draft tokens. It builds a Trie (prefix tree) from retrieved continuations, where node weights reflect token sequence frequencies. Speculative RAG~\cite{wang2024speculative} use a fine-tuned specialist LM to generate complete answer drafts with supporting rationales. It clusters retrieved documents by similarity, generates diverse drafts from different document subsets, and employs self-consistency and self-reflection scores for draft evaluation instead of token-level verification.
\subsection{N-gram-based Methods}
\label{sec:ngram_methods}
Several approaches leverage n-gram patterns for efficient token generation. ANPD~\cite{ou2024lossless} replaces traditional draft models with an adaptive N-gram system that updates predictions based on context. LOOKAHEAD~\cite{fu2024break} uses n-gram verification by collecting and utilizing n-grams from previous iterations as draft tokens. The N-Grammys~\cite{stewart2024n} further develops this idea by creating a dedicated n-gram based prediction system that can operate without requiring a separate draft model.

\subsection{Auto-regressive Generation}
\label{sec:auto_regressive}
Most sequence generation methods employ auto-regressive drafting, where a smaller model generates draft tokens that are verified by a larger target model. This drafting paradigm has spawned numerous techniques that vary in how the draft model interacts with the target model.

\subsubsection{Independent Drafters}
\label{sec:independent_drafter}
Auto-regressive independent drafters are techniques in which smaller model(s) generate tokens one at a time while a separate larger target model subsequently verifies the draft tokens in parallel. SpecDec~\cite{xia2023speculative} pioneered this approach with an independent draft model using distinct attention queries for masked positions. SpecDec++~\cite{huang2024specdec++} improves SpecDec~\cite{xia2023speculative} by training a prediction head on top of the draft model that estimates the probability of token acceptance by the target model. Based on these predictions, it dynamically determines when to stop generating tokens and trigger verification.

 Recent works focus on dynamic adaptation and confidence monitoring. BiLD~\cite{kim2024speculative} triggers target model verification when draft confidence falls below a threshold, while ON-THE-FLY~\cite{liu2025a} dynamically adjusts window sizes based on prediction accuracy. OSD~\cite{liu2023online} enables online adaptation through knowledge distillation during inference, and DistillSpec~\cite{zhou2023distillspec} extends this by accessing target model logits for improved alignment.
\cite{Liu2025} introduces special tokens for draft models to autonomously determine target model consultation, eliminating separate verification at some performance cost. For mathematical applications, Judge\cite{bachmann2025judge} adds a learned verification layer atop the target model's embeddings, using contextual correctness assessment to reduce strict output alignment requirements.

\subsubsection{Dependent Drafters}
\label{sec:dependent_drafter}
The main drawbacks of independent drafting approaches are that  (1) the computation required to generate the draft tokens is fixed per tokens, meaning that computation is over-provisioned for many ``easy'' tokens and (2) the target model cannot reuse the features of the drafting process, increasing the amount of compute required. Self-speculative decoding approaches generate draft tokens by relying directly on a subset (\textbf{Layer Skipping}) or extension (\textbf{Dependent Heads}) of the target model.

\paragraph{Layer Skipping} Draft\&Verify~\cite{zhang2023draft}, SWIFT~\cite{xia2024swift}, and Draft on the Fly~\cite{metel2024draft} achieves fast draft token generation by selectively skipping some intermediate layers in the Draft process, and then verifies these drafts using the full LLM. In order to achieve good draft accuracy, they also designed an intermediate layer selection algorithm based on Bayesian optimization. LayerSkip~\cite{elhoushi2024layer} uses an early exiting~\cite{teerapittayanon2016branchynet} approach to dynamically output tokens at different depths of the target model. Kangaroo~\cite{liu2024kangaroo} also applied early exit by adopting a shallow sub-network to generate drafts and using a lightweight adapter module to bridge the performance gap with the full model, achieving efficient and accurate decoding. EESD~\cite{liu2024speculative} use Thompson Sampling Control~\cite{slivkins2019introduction} Mechanism to adaptively determines how many draft token will be generated. SPEED~\cite{hooper2023speed} combines speculative execution with parameter sharing, using early predictions to process multiple tokens in parallel through shared decoder layers, rather than waiting for each token to complete sequentially.

\paragraph{Dependent Heads}
Dependent head-based drafting eliminates the need for a separate draft model by adding lightweight feed-forward prediction heads using the hidden states of the target model. The main idea is that the first token in sequence generation block uses the target model as usual but the features at the end of the model are fed into additional heads to predict subsequent tokens without passing back through the entire target model.

EAGLE~\cite{li2024eagle} uses a trained head that takes in hidden states from the target model and generates subsequent draft tokens in an AR manner. Hydra~\cite{ankner2024hydra} use multiple decoding, one for each draft token position. 

EAGLE extensions have focused on improving parallel token generation and attention mechanisms. Falcon~\cite{gao2024falcon} introduces a semi-autoregressive framework combining LSTM layers and relaxed causal-masked self-attention to generate k tokens per forward pass, while HASS~\cite{zhang2024learning} enhances knowledge distillation by prioritizing high-probability tokens during training. Mixture of Attentions~\cite{zimmer2024mixture} incorporates multiple attention types (LSA, SA, and CA) for improved token prediction, and DeepSeek-V3~\cite{liu2024deepseek} adapts ~\cite{gloeckle2024better}'s multi-token approach (discussed in Section~\ref{sec:multi_token_pre}) while maintaining complete causal attention during inference.

\subsection{Multi-token Prediction}
\label{sec:multi_token_pre}

\cite{stern2018blockwise} proposes adding multiple decoding heads on top of a model to predict $k$ future tokens in parallel, requiring training the entire model from scratch. Medusa~\cite{cai2024medusa} introduces a parameter-efficient approach, where lightweight decoding heads are fine-tuned on top of pre-trained language models. Each head is trained to predict a specific future position in the sequence without modifying the target model. ~\cite{gloeckle2024better} propose a multi-token prediction paradigm where a shared backbone optimized jointly with multiple prediction heads that enable propagation of information related to sequential tokens during training that can be discarded at inference to enable parallel generation (similar to Medusa). 

Recent improvements enhance Medusa's independent draft heads by modeling inter-token relationships. Amphista~\cite{li2024amphista} uses bi-directional self-attention to consider both past and future predictions, while CTC Drafting~\cite{wen2024speculative} employs Connectionist Temporal Classification (CTC) with blank tokens and repetition, followed by duplicate removal to generate draft sequences.

\section{Sequence Refinement Methods}
\label{sec:refine}

\subsection{Single-pass Verification}
\label{sec:refine:singlepass}

Single-pass verification represents the most common refinement strategy in draft-and-verify approaches, where drafted tokens are verified exactly once by the target model. 

\subsubsection{Linear Verification}
\label{sec:refine:linear}
Linear verification sequentially validates draft tokens against the target model's logit distributions, with early works like SpecDec~\cite{xia2023speculative} and Draft\&Verify~\cite{zhang2023draft} comparing drafted tokens against the target model's predictions. When a token fails verification (i.e., when the draft output doesn't match the target model's distribution), the system falls back to standard AR generation from that point.

Fast Inference ~\cite{leviathan2023fast} and ~\cite{chen2023accelerating} introduced speculative sampling to improve acceptance rates while approximately maintaining the target distribution. Their method accepts a token if the target model assigns equal or higher probability; otherwise, it accepts with probability $p(x)/q(x)$ or resamples from an adjusted distribution.

Block Verification~\cite{sun2025block} and MTAD~\cite{qin2024optimized} improve upon linear verification by examining the joint probability distribution of draft tokens as a chain of conditional probabilities. This block-based evaluation approach typically results in higher acceptance rates compared to token-by-token verification for similar quality.

\subsubsection{Tree-based Verification}
\label{sec:refine:tree_based}
Tree-based verification extends the single-pass paradigm by enabling parallel exploration of multiple completion paths. Unlike linear verification that processes a single sequence, tree-based methods construct and verify a tree of possible completions simultaneously, making more efficient use of parallel compute resources.


SpecInfer~\cite{miao2023specinfer} pioneered this approach by developing an efficient tree-based attention masking scheme that enables parallel verification while maintaining proper token dependencies. This innovation maintains generation quality while significantly increasing the number of tokens that can be verified in parallel.

Recent works have focused on optimizing tree structure and size to maximize computational efficiency. Sequoia~\cite{chen2024sequoia} introduces a hardware-aware tree optimizer that can maximize inference performance by selecting appropriate tree dimensions based on available computing resources. OPT-Tree~\cite{wang2024opt} searches for optimal tree structures to maximize expected acceptance length per decoding step. DSBD~\cite{qin2024dynamic} uses a small model to generate multiple candidate sequences via beam search, then the large model verifies these sequences layer by layer while dynamically adjusting the beam width based on acceptance probabilities to balance efficiency and quality. DySpec~\cite{xiong2024dyspec} enables dynamic tree expansion during runtime based on prediction confidence, while EAGLE2~\cite{li2024eagle2fasterinferencelanguage} incorporates context-aware tree construction to improve acceptance rates. DDD~\cite{brown2024dynamic} optimizes EAGLE2~\cite{li2024eagle2fasterinferencelanguage} 's tree drafting method by making the depth dynamic based on draft model confidence.

Several works have explored hybrid approaches that combine tree-based verification with other techniques. ProPD~\cite{zhong2024propd} integrates progressive refinement into the tree structure, while RSD~\cite{jeon2024recursive} employs recursive verification strategies. GSD~\cite{gong2024graph} and ADED~\cite{liu2024adaptive} extend tree-based methods to handle more complex dependency structures through graph-based representations and adaptive depth adjustment.

In terms of verifying multiple candidate draft tokens in parallel (also known as Multi-Draft Speculative Decoding, MDSD), ~\cite{hu2025towards} propose a hybrid sampling strategy that combines deterministic selection of high-probability tokens with random sampling of the final token, improving acceptance rates in certain scenarios.  ~\cite{khisti2024multi} introduce a two-phase verification method that uses importance sampling to select a draft token before applying single-draft verification, optimizing the process for parallel draft generation.

\begin{figure}
    \centering
    \includegraphics[width=1\linewidth]{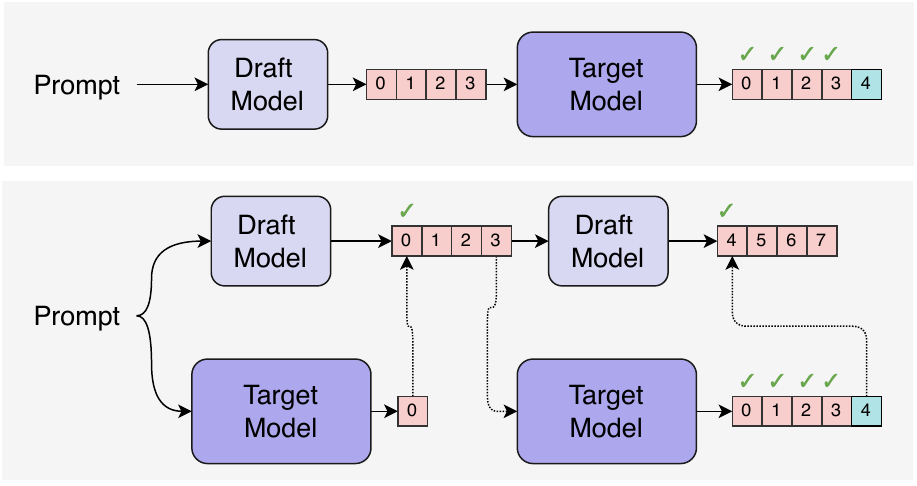}
    \caption{Comparison of speculative decoding approaches: (a) Sequential processing where draft generates tokens (0-3) before target verification. (b) Parallel processing where draft generates new tokens while target simultaneously verifies previous ones.}
    \label{fig:parallel-sd}
\end{figure}

\subsection{Iterative Decoding}
\label{sec:refine:iterative}
Iterative decoding methods extend the single-pass verification paradigm by allowing multiple refinement iterations on draft tokens until convergence. These approaches draw inspiration from classical numerical methods for solving systems of nonlinear equations, particularly the Jacobi and Gauss-Seidel iteration methods.

In ~\cite{santilli2023accelerating}, the authors reframe AR text generation as an iterative optimization problem. Their approach expresses token generation as a system where each position must output the most likely token given the current state of all other positions. Starting with a randomly initialized sequence, they adapt the Jacobi method to update all positions in parallel during each iteration until convergence. The authors prove that this process produces identical output to traditional AR decoding under greedy sampling. ~\cite{fu2024break} builds upon this framework with LOOKAHEAD decoding, which combines Jacobi iterations with n-gram verification to accelerate convergence by leveraging predictions from earlier steps.

 CLLMs~\cite{kou2024cllms} leverages consistency training to accelerate convergence by enabling better multi-token prediction in early iterations. 
\section{System-Level Optimizations and Implementation Strategies}
\label{sec:system-sd}

\begin{figure}
    \centering
    \includegraphics[width=1\linewidth]{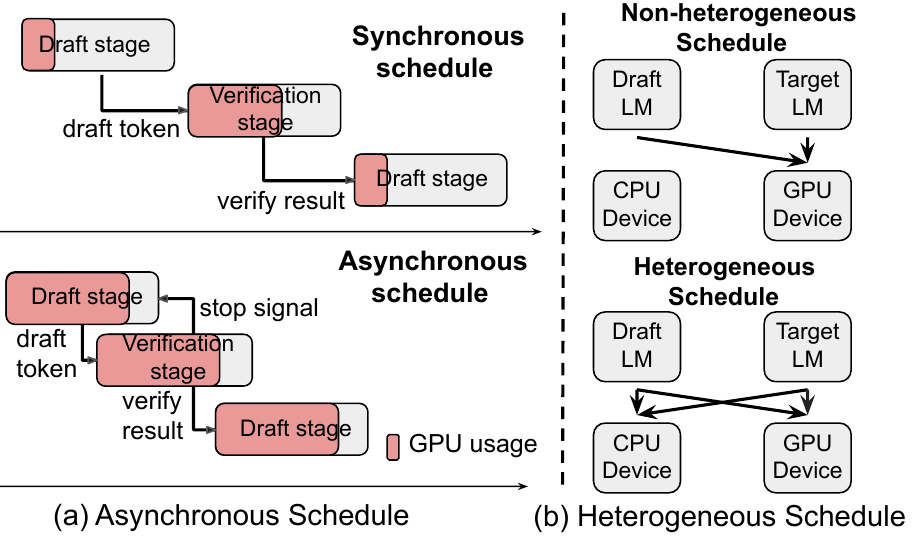}
    \caption{Asynchronous and heterogeneous schedules.}
    \label{fig:sys-sd}
\end{figure}

\subsection{Parallel Speculative Decoding}
\label{sec:parallel-sd}

Traditional SD processes tokens sequentially, with the draft model generating tokens followed by target model verification, creating inherent bottlenecks. As shown in Figure~\ref{fig:parallel-sd}, parallel approaches overcome this limitation by enabling simultaneous operation - while the target model verifies earlier tokens, the draft model generates subsequent ones, enabling continuous overlapped execution. Recent methods build upon this paradigm: CS Drafting~\cite{chen2023cascade} employs vertical and horizontal cascade structures for 81\% speedup, PaSS~\cite{monea2023pass} uses look-ahead embeddings for 30\% speedup, and Faster Cascades~\cite{narasimhan2024faster} incorporates deferral rules for improved cost-quality trade-offs. PEARL~\cite{liu2024parallel} further advances this through pre-verify and post-verify strategies with adaptive draft lengths, achieving 4.43$\times$ speedup over AR decoding and 1.50$\times$ over standard SD AMUSD~\cite{mcdanel2024amusd} presents an asynchronous multi-device approach to SD, decoupling the draft and verify phases into continuous, asynchronous operations.

\subsection{Distributed Speculative Decoding}
\label{sec:edge-sd}
Edge computing environments impose stringent constraints on memory, compute power, and latency, necessitating specialized SD approaches to deploy LLMs effectively in resource-constrained settings. SpecExec~\cite{svirschevski2024specexec} is designed to harness the parallel processing power of consumer GPUs to accelerate LLM inference. By generating multiple tokens per target model iteration and constructing a ``cache'' tree of probable continuations, SpecExec efficiently validates these continuations with the target model in a single pass. EdgeLLM~\cite{xu2024edgellm} further optimizes on-device LLM inference through novel techniques for resource allocation and error correction, achieving great token generation speeds and significantly outperforming existing engines. Dovetail~\cite{zhang2024dovetail} represents a significant advancement in heterogeneous computing for LLM inference. By deploying the draft model on the GPU and the target model on the CPU, Dovetail reduces the granularity of data transfer and enhances the overall inference process. The introduction of Dynamic Gating Fusion (DGF) and optimizations for low-end hardware further improve the balance between latency and performance.

\subsection{Compiler and Hardware Optimization for Speculative Decoding}
\label{sec:batch-sd}
Efficient implementation of SD requires careful optimization of both hardware resources and compiler strategies to maximize throughput and minimize latency. SpecPIM~\cite{li2024specpim} presents a novel approach to accelerate speculative inference on a Processing-in-Memory (PIM) system through co-exploration of architecture and dataflow. This method constructs a design space that comprehensively considers algorithmic and architectural heterogeneity, enabling optimal hardware resource allocation for different models and computational patterns. ~\cite{wagner2024optimized} investigates improvements in speculative sampling on GPUs, achieving significant speed gains by parallelizing computations and using sigmoid approximations for softmax, though this comes with a minor reduction in accuracy.

\begin{figure}
    \centering
    \includegraphics[width=0.75\linewidth]{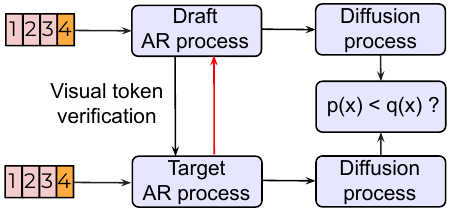}
    \caption{Flow of AR image generation with SD.}
    \label{fig:ar-mutimodal}
\end{figure}

Recent studies have focused on enhancing the throughput of LLMs using SD by optimizing batch processing and scheduling strategies. Figure~\ref{fig:sys-sd} illustrates two scheduling strategies for SD systems: (a) Asynchronous Schedule: The draft stage is followed by the verify stage, with optional stop signals determining further processing. This non-blocking approach enhances system efficiency. (b) Heterogeneous Schedule: Both CPU and GPU devices are utilized for different stages of the decoding process, enabling parallel processing and optimizing performance through resource allocation. Using Markov chain theory, \cite{yin2024theoretical} establishes SD's optimality among unbiased algorithms while highlighting the tradeoff between inference speed and output quality. Their analysis reveals that batch processing benefits are limited by the distribution gap between small and large models. MagicDec~\cite{chen2024magicdec} identifies the shift from compute-bound to memory-bound bottlenecks as batch size and sequence length increase, using sparse KV caches in draft models to optimize throughput. BASS~\cite{qian2024bass} extends SD to a batched setting with customized CUDA kernels for ragged tensors in attention calculations and dynamically adjusts draft lengths for better GPU utilization. SEED~\cite{wang2024seed} accelerates reasoning tree construction through scheduled speculative execution, using a rounds-scheduled strategy for conflict-free parallel processing. PipeInfer~\cite{butler2024pipeinfer} addresses single-request latency through pipelined speculative acceleration, reducing inter-token latency via asynchronous speculation and early cancellation. TRIFORCE~\cite{sun2024triforce} introduces a hierarchical SD mechanism with a dynamic sparse KV cache to achieve lossless acceleration of long sequence generation, significantly improving generation speed and efficiency while maintaining quality. ~\cite{zhao2024qspec} proposes QSPEC, a novel framework that combines weight-shared quantization schemes with SD, achieving up to 1.55× acceleration without quality loss, paving the way for efficient and high-fidelity quantization deployment in diverse and memory-constrained settings. ~\cite{wang2024mamba} introduces a hardware-aware SD algorithm that accelerates the inference speed of Mamba and hybrid models. Inspired by SD, SKD~\cite{xu2024speculative} represents a novel, adaptive approach to knowledge distillation. By dynamically generating tokens and using the teacher model to filter or replace low-quality samples, it bridges the gap between supervised KD's reliance on static data and on-policy KD's susceptibility to low-quality outputs. This ensures a better alignment between training and inference distributions, and improved performance.
\section{Multimodal Models and Applications}
\label{sec:sd_application}

\subsection{Speculative Decoding for Visual Output Generation}
\label{sec:AR_visual}
Researchers are now using SD to improve the efficiency of AR image generation~\cite{ding2021cogview,yu2022scaling, li2024autoregressive}. As shown in Figure~\ref{fig:ar-mutimodal}, this method greatly speeds up the process by reducing the inference steps needed for generating visual tokens.
For instance,~\cite{wang2024continuous} proposes a novel continuous SD method that designs a novel acceptance criterion for the diffusion distributions, significantly improving the efficiency of AR image generation. Similarly, LANTERN~\cite{jang2024lantern} presents a relaxed acceptance condition for the SD strategy to substantially speed up the inference process in visual AR models. Additionally, Speculative Jacobi Decoding (SJD)~\cite{teng2024accelerating} offers a training-free speculative Jacobi decoding technique that effectively accelerates text-to-image generation tasks.

\subsection{Speculative Decoding for Multimodal Output Generation}
\label{sec:mutimodel}

Recent advancements in SD have substantially improve the efficiency and quality of AR generation across various modalities. In the domain of speech synthesis, VADUSA~\cite{li2024fast} leverages SD to accelerate the inference process in AR text-to-speech (TTS) systems, which enhances the quality speech synthesis as well. Inspired by the flavor of SD, ~\cite{raj2024faster} introduces a multi-token prediction mechanism, offering substantial improvements in inference efficiency for speech generation.

In the context of multimodal large language models, ~\cite{gagrani2024speculative} investigates the integration of SD into the LLaVA 7B model to optimize inference efficiency. Their findings indicate that employing a lightweight, language-only draft model facilitates a memory-constrained acceleration of up to 2.37×. Besides, IbED~\cite{leebatch} proposes the "In-batch Ensemble Drafting" method to further enhance the robustness and efficiency of SD. It adopts the ensemble techniques during batch-level inference, requires no additional model parameters and significantly increases the validation probability of draft tokens, thereby improving performance and robustness across diverse input scenarios.

\subsection{Recommendation Systems}
\label{sec:SR_apps}
LLM-based recommendation systems have shown great potential in enhancing personalized recommendations, but their high inference latency poses a significant challenge for real-world deployment. To address this, recent research has focused on optimizing decoding efficiency to accelerate recommendation generation. ~\cite{xi2024decoding} propose DARE that integrates retrieval-based SD to accelerate recommendation knowledge generation, thereby improving the deployment efficiency of LLM-based recommender systems in industrial settings. AtSpeed~\cite{lin2024efficient} combines strict top-K alignment (AtSpeed-S) and relaxed sampling verification (AtSpeed-R), to significantly accelerate LLM-based generative recommendation with speedup from 2$\times$ to 2.5$\times$, addressing inference latency challenges in top-K sequence generation.
\section{Conclusion}
\label{sec:conslusion}
This survey analyzes generation-refinement frameworks for mitigating sequential dependencies in autoregressive models, highlighting how these approaches are fundamentally changing efficient neural sequence generation across text, speech, and visual domains. Through examining both algorithmic innovations and system-level implementations, we have demonstrated their broad applicability while providing crucial deployment insights for practitioners. Moving forward, significant challenges persist in constructing solid theoretical foundations to grasp the balance between parallelism and quality, as well as in developing comprehensive approaches that span different modalities—efforts that could narrow the divide between the capabilities of large models and their actual implementation. Additionally, it remains crucial to examine the scalability of the speculative decoding system as the quantity of draft and target models increases.


\bibliographystyle{IEEEtran}
\bibliography{refs}

\end{document}